\documentclass[onecolumn, a4paper,journal,12pt]{IEEEtran}
\usepackage{amsfonts}
\usepackage{graphicx, epsfig,amsmath,amssymb,latexsym,graphics,float}
\usepackage{setspace,subfig}
\usepackage{citesort}

\include{IEEEtran.bib}

\begin{document}

\title{``Weak AI" is Likely to Never Become ``Strong AI", So What is its Greatest Value for us?}
\author{$^{\star}$Bin~Liu\\
First posted March 30th, 2021\\
\thanks{$^{\star}$B. Liu is with Zhejiang Lab, Hangzhou, China. e-mail: bins@ieee.org or liubin@zhejianglab.com.}
}
\maketitle
\begin{abstract}
AI has surpassed humans across a variety of tasks such as image classification, playing games (e.g., go, ``Starcraft" and poker), and protein structure prediction. However, at the same time, AI is also bearing serious controversies. Many researchers argue that little substantial progress has been made for AI in recent decades. In this paper, the author (1) explains why controversies about AI exist; (2) discriminates two paradigms of AI research, termed ``weak AI" and ``strong AI" (a.k.a. artificial general intelligence); (3) clarifies how to judge which paradigm a research work should be classified into; (4) discusses what is the greatest value of ``weak AI" if it has no chance to develop into ``strong AI".
\end{abstract}

\section{Introduction}\label{sec:intro}
The last decade has seen impressive applications of AI represented mostly by deep neural networks, i.e., deep learning \cite{lecun2015deep}. The striking point lies in that the computing agent has reached and even surpassed humans in many tasks, e.g., image classification \cite{krizhevsky2012imagenet}, speech recognition \cite{xiong2018microsoft,saon2017english}, games \cite{silver2018general,brown2018superhuman,vinyals2019grandmaster},  protein structure prediction \cite{senior2020improved}. Even ten years ago, it was hard to imagine that AI would achieve so many amazing breakthroughs.

On the other side, AI is also bearing serious controversies during the same period. Among the critics, Judea Pearl, a pioneer for probabilistic reasoning in AI and a winner of the Turing award, argues that ``... \emph{all the impressive achievements of deep learning amount to just curve fitting}," and a necessary ability to be supplemented for AI is causal reasoning \cite{pearl2019limitations,pearl2018ai}.  Gary Marcus, a professor of cognitive science, summarizes ten limitations of deep learning \cite{marcus2018deep}, namely, ``\emph{... it is data-hungry, ... it has limited capacity for transfer, ... it has no natural way to deal with
hierarchical structure, ... it struggles with open-ended inference, ... it is not sufficiently transparent, ... it has not been well integrated with prior knowledge, ... it cannot inherently distinguish causation
from correlation, ... it presumes a largely stable world, in ways that may
be problematic, ... it works well as an approximation, but its
answers often cannot be fully trusted, ... it is difficult to engineer with}". In a recent issue of the journal Frontiers in Psychology, another cognitive scientist J. Mark Bishop argues that AI ``\emph{is stupid and causal reasoning will not fix it}" \cite{bishop2021artificial}.

In this paper, I attempt to concisely respond to current controversies about AI. Specifically, I emphasize discrimination between two paradigms of AI research, namely ``weak AI" and ``strong AI" (Section \ref{sec:two_AI}); provide a conceptual guide to judge which paradigm a research work should be classified into (Section \ref{sec:judge}), explain why controversies about AI last (Section \ref{sec:controversies}), present major views on whether ``weak AI" will grow into ``strong AI" (Section \ref{sec:will}) and discuss what is the greatest value of ``weak AI" if it has no chance to become ``strong AI" (Section \ref{sec:value}).
\section{What do ``Weak AI" and ``Strong AI" Mean?}\label{sec:two_AI}
``Weak AI" and ``Strong AI" are two terms coined by John Searle in the ``Chinese room argument" (CRA) \cite{john1980minds}. CRA is a thought experiment as follows: ``\emph{Searle imagines himself alone in a room following a computer program for responding to Chinese characters slipped under the door. Searle understands nothing of Chinese, and yet, by following the program for manipulating symbols and numerals just as a computer does, he sends appropriate strings of Chinese characters back out under the door, and this leads those outside to mistakenly suppose there is a Chinese speaker in the room}" \cite{crm}. The term ``strong AI" entails that, ``\emph{... the computer is not merely a tool in the study of the mind; rather, the appropriately programmed computer really is a mind, in the sense that computers
given the right programs can be literally said to understand and have other cognitive
states.}" In contrast, the term ``weak AI" implies that ``\emph{... the principal value of
the computer in the study of the mind is that it gives us a very powerful tool.}"
J. Mark Bishop summarizes that `\emph{`weak AI focuses on epistemic issues relating to engineering a simulation of human intelligent behavior, whereas strong AI, in seeking to engineer a computational
system with all the causal power of a mind, focuses on the ontological} " \cite{bishop2021artificial}.

I borrow the terms ``weak AI" and ``strong AI" here without an intent to discuss CRA. See related discussions in e.g., \cite{Rey1986,shaffer2009logical,sloman1980turn,boden1988computer}.

Simply put, ``weak AI" represents computational systems that exhibit as if they own human intelligence, but they do not. In contrast, ``strong AI" represents computational systems that have human intelligence. Correspondingly, all AI research can be categorized into two paradigms: one is targeted for realizing ``strong AI"; and the other produces advanced ``weak AI" systems to meet a variety of practical needs.
\subsection{How to Judge a Research Work Belongs to Which Paradigm?}\label{sec:judge}
The biggest motivation for realizing ``strong AI" is to answer the question: what are the generation mechanisms of humans' intelligence and how to implement these mechanisms with a machine. Therefore, given a research work, it is easy to judge whether it belongs to the ``strong AI" paradigm. If this work provides any new and useful clue for us to answer the above question, it falls within the ``strong AI" paradigm; otherwise, it belongs to the ``weak AI" paradigm.

Based on the above method, part of the (especially early) works on neural networks that deepen our understanding of the working mechanism of biological neural systems, surely belongs to the ``strong AI" paradigm. On the other hand, most research works that involve artificial neural networks and deep learning, even if they are proposed under the inspiration of research on neuroscience, cognitive science, behavior psychology, they belong to the ``weak AI" paradigm as long as they do not give us any new insight on the generation mechanisms of humans' intelligence or on how to better implement mechanisms that have already been found.
\section{Why Controversies about AI Last?}\label{sec:controversies}
In controversies about AI, party A believes that AI has made substantial progress in the past decade;  party B doubts or even negates the development of AI.

I argue that controversies arise mainly because these two parties mix two different concepts, ``weak AI" and ``strong AI", together, when they talk about AI. The fact is that ``weak AI" has made substantial progress in the past decade, while ``strong AI" has not. Party A thinks that ``weak AI" is an important member of the AI family; progress gained from ``weak AI" also belongs to this AI family. In contrast, in the mind of Party B, there always exists one ideal form of AI, namely a realized ``strong AI", and the ``distance" between current AI and this ideal AI is treated as a criterion for evaluating current AI. Compared with decades ago, current AI still lacks basic human-level abilities such as causal reasoning \cite{pearl2019limitations}, robust decision making \cite{dietterich2017steps}, commonsense utilization \cite{marcus2020next}, and knowledge transfer, which implies that the ``distance" between the realized AI and the ideal ``strong AI" has not been remarkably shortened. Therefore, it is reasonable for party B to doubt or even negate the development of AI.
%

A natural question arises: how breakthroughs of ``weak AI" have come out in the past decade?
Judea Pearl argues that ``... \emph{all the impressive achievements of deep learning amount to just curve fitting"}. However, the point is that, different from previous fitting methods, deep learning permits to do an extraordinary fitting - fitting multi-modal big data in an end-to-end way. This deep learning type of fitting requires a big consumption of both computing and storage resources but avoids labor-intensive feature engineering. Big data, big computing, and big storage are three requisites that make deep learning surpass humans in playing Go, image classification, speech recognition, and so on. The luck for deep learning is that the past decade happens to witness great improvements in sensing technologies, wireless mobile phones, cloud computing, computing devices, computer storage, and databases, which give birth to big data, big computing, and big storage required by deep learning.
%
%
\section{Will ``Weak AI" Grow into ``Strong AI"?}\label{sec:will}
A metaphor is often used to reply to this question: the relationship between ``weak AI" and ``strong AI" is like that between flying machines and birds. Flying machines are not developed by accurately mimicking birds' flying. Birds perform much better in maneuvering than the most advanced flying machine today. Birds can flexibly re-purpose their behaviors while flying machines cannot. But the appearance of flying machines has met demands of speedy transportation and others. People may think that, since it is unlikely and not necessary for flying machines to develop into birds, then similarly, ``weak AI" is unlikely and not necessary to grow into ``strong AI".

To formally consider whether ``weak AI" will grow into ``strong AI", let recall the Turing test \cite{turing2009computing} and the CRA (mentioned in Section \ref{sec:two_AI}). An example statement of the Turing test is as follows \cite{ibm}: ``\emph{Originally known as the Imitation Game, the test evaluates if a machine’s behavior can be distinguished from a human. In this test, there is a person known as the “interrogator” who seeks to identify a difference between computer-generated output and human-generated ones through a series of questions. If the interrogator cannot reliably discern the machines from human subjects, the machine passes the test. However, if the evaluator can identify the human responses correctly, then this eliminates the machine from being categorized as intelligent.}" Through the lens of CRA, Searle argues that the Turing test has serious flaws, as passing the test does not indicate that the machine has consciousness or understanding. The absence of an effective evaluation method hampers the development of ``strong AI".

Besides, philosophers and cognitive scientists often use G$\ddot{\mbox{o}}$del's first incompleteness theorem \cite{godel} to argue that a machine cannot generate humans' consciousness or understanding. See related discussions in e.g., \cite{bishop2021artificial}.
\section{What is the Greatest Value of ``Weak AI" for us?}\label{sec:value}
In his most recent paper, Geoffrey Hinton states that ``\emph{The difference between science and philosophy is that experiments can show that extremely plausible ideas are just wrong and extremely implausible ones, like learning an entire complicated system by end-to-end gradient decent, are just right}" \cite{hinton2021represent}. In \cite{pearl2020radical}, Judea Pearl argues that ``\emph{Modern connectionism has in fact been viewed as a Triumph of Radical Empiricism over its rationalistic rivals. Indeed, the ability to emulate knowledge acquisition processes on digital machines offer enormously flexible testing grounds in which philosophical theories about the balance between empiricism and innateness can be submitted to experimental evaluation on digital machines.}" Combining their arguments, one can see that they both attribute recent deep learning's success as a success of empiricism which is data-driven, other than driven by philosophical theory or intuition.

A very important lesson that can be learned from the fast-pacing development and applications of AI in the past decade is that deep learning running on big enough data can produce unexpected shortcuts to solve extremely difficult problems. For example, by combining deep learning, reinforcement learning \cite{sutton2018reinforcement}, and Monte Carlo tree search \cite{gelly2011monte}, a computer program AlphaGO \cite{silver2016mastering} can win the human champion without having to understand any of the Go-playing strategies that have been accumulated by humans for more than four thousand years. The Generative Pre-trained Transformer 3 (GPT-3) \cite{brown2020language} can generate human-like texts through deep learning without having to understand any syntax or semantics underlying the texts.
It is shown that the greatest value of ``weak AI" represented by deep learning lies in that it provides scalable, less-labor-involved, accurate, and generalizable tools for distilling, representing and then exploiting patterns hidden from big data. Although such ``weak AI" has no real intelligence, to a large extent it meets urgent needs for scalable, efficient, and accurate processing of big data.

In a foreseeable future, ``weak AI" is likely to become more robustly (with e.g., portfolio  \cite{dietterich2017steps} or dynamic portfolio methods \cite{liu2020sequential,qi2019dynamic,liu2017robust,dai2016robust,liu2020data,liu2011instantaneous}), while a big challenge is how to model ``unknown unknowns"; it will perform more automatically through e.g., auto machine learning \cite{hutter2019automated}, but it can not become completely automatic provided that ``strong AI" is realized \cite{liu2018very}; it may perform as if it owns abilities of cognition and understanding, but it does not.
\section{Conclusions}
AI has made great progress in the past decade. It has influenced almost all facets of human society by providing more efficient algorithmic solutions to representation, management, analysis of multi-modal big data. Controversies about AI last mainly because ``weak AI" becomes so strong while ``strong AI" is almost as weak as it was decades ago. Almost all breakthroughs of AI that have attracted the public's attention in the past decade are within the ``weak AI" paradigm. ``Weak AI" is developing much faster than expected. Even ten years ago, one could not imagine that a computer program would beat the human champion soon in playing Go. In contrast, the ``fruits" people have got from the ``strong AI" paradigm are not so striking as from ``weak AI". I suggest, when talking about AI in the future, one should better make a statement in advance whether this talk is about ``weak AI" or ``strong AI". In this way, more focused and constructive discussions can be expected.

In a foreseeable future, ``weak AI" cannot develop into ``strong AI" (see why in Section \ref{sec:will}), but it provides a channel to synthesize advances obtained from related disciplines such as cloud computing, computer storage, high-speed wireless mobile communications. Through this synthesis of technologies, more advanced algorithmic tools will be developed in the ``weak AI" paradigm, then ``weak AI" will continue to influence human society more profoundly, through big data. The man-computer symbiosis world that Licklider predicted more than sixty years ago \cite{licklider1960man} is becoming a reality.
\bibliographystyle{IEEEtran}
\bibliography{mybibfile}
\end{document}